\pdfoutput=1

\documentclass[11pt]{article}

\usepackage[]{emnlp2021}

\usepackage{times}
\usepackage{latexsym}

\usepackage{booktabs}
\usepackage{todonotes}
\usepackage{latexsym}
\usepackage{amsmath}
\usepackage{bbm}
\usepackage{amssymb}

\usepackage{multirow}
\usepackage{makecell}
\usepackage{subcaption}

\usepackage[T1]{fontenc}

\usepackage[utf8]{inputenc}

\usepackage{microtype}

\title{Are you doing what I say? \\ On modalities alignment in ALFRED}

\author{Ting-Rui Chiang \thanks{\ \ Chiang and Yeh have equal contributions} \and Yi-Ting Yeh\footnotemark[1] \and Ta-Chung Chi \and Yau-Shian Wang \\
        Carnegie Mellon University  \\ 
        \texttt{\{tingruic,yitingye,tachungc,yaushiaw\}@andrew.cmu.edu} }

\begin{document}
\maketitle
\begin{abstract}
ALFRED is a recently proposed benchmark that requires a model to complete tasks in simulated house environments specified by instructions in natural language.
We hypothesize that key to success is accurately aligning the text modality with visual inputs.
Motivated by this, we inspect how well existing models can align these modalities using our proposed intrinsic metric, boundary adherence score (BAS).
The results show the previous models are indeed failing to perform proper alignment. 
To address this issue, we introduce approaches aimed at improving model alignment and demonstrate how improved alignment, improves end task performance.
\end{abstract}

\section{Introduction and Problem Definition}

Recently several datasets have been proposed to benchmark machine's capability of following instructions in simulated real world environments \cite{anderson2018vision,krantz2020beyond,wu2018building,hermann2020learning,misra2018mapping,chen2019touchdown,shridhar2020alfred}.
To accomplish these tasks well, an agent needs to map the instruction given in natural language to actions while interacting with the environment simulator.
ALFRED~\cite{shridhar2020alfred} is one of them that requires complex interaction.
Each task in ALFRED involves a long sequence of instructions.
To carry out the given task, an agent needs to navigate through the environment and interact with objects by generating pixel-level masks.
It is thus very similar to real world scenarios.

Given this task, we ask a question: "\textit{To what degree can a model align the literal instructions with its real world interactions?}
We hypothesize this to be key to success. 
As human beings, when we are following a sequence of instructions, we know what the instruction we are working on, so we know what to do, and what to do next after finishing this instruction.
For this reason, we conjecture such alignment is also necessary for a model.

Motivated by our hypothesis, we first investigate the existing models' capability of aligning the literal instructions and the visual inputs it perceives.
To quantify this capability, we propose an intrinsic metric Boundary Adhere Score (BAS).
It measures how frequently the model is focusing on the corresponding step-by-step instruction when predicting an action.
We inspect the two publicly available baselines, the Seq2Seq model proposed in~\cite{shridhar2020alfred} and MOCA~\cite{pratap2020moca}.
The results indicate that both of the two have sub-optimal alignment.

Therefore, we propose methods to improve the alignments from two aspects.
1) We propose a novel neural program counter model, which explicitly keeps track of the instruction that is being executed.
2) We propose a new auxiliary loss $L_{pc}$ that provides stronger learning signal for learning the alignment.

Our preliminary results show that the program counter module along with the auxiliary loss are promising.
It outperforms the original state-of-the-art model.
We also apply our BAS metric on the MOCA model with our proposed program counter and auxiliary loss.
The results show that they can indeed improve the alignment.
It is aligned with our hypothesis that alignment is important.

To sum up, our contribution is three-fold:
1) We propose an intrinsic metric that quantifies the alignment.
2) Our analysis indicates that the previous models do not align the modalities very well.
3) Based on our analysis we propose improvement over the current state-of-the-art model, leading to a promising future research direction.

\section{Related Work}

\subsection{Vision and Language Navigation Tasks}
Vision and language navigation tasks bear a strong resemblance to ALFRED, as they require an agent to achieve a goal by following instructions in a simulated environment.

\paragraph{R2R}
Stanford large-scale 3D Indoor Spaces~\cite{3D} and Room2Room (R2R)~\cite{anderson2018vision} are two benchmarks for Vision-and-Language Navigation (VLN).
In R2R, the whole system is built upon the Matterport-3D simulator, which encodes indoor environments as graphs where nodes are navigable waypoints and edges represent the feasibility to go from one waypoint to another.

\paragraph{VLN-CE}
\citeauthor{krantz2020beyond,wu2018building} further proposes Vision-and-Language Navigation in Continuous Environments (VLN-CE) which doesn't provide pre-defined navigation graphs to an agent. An agent needs to navigate via egocentric visual inputs as opposed to panoramic viewpoints.

\paragraph{StreetNav}
Other than indoor environments, StreetNav \cite{hermann2020learning} additionally provides thumbnails to an agent to navigate in Google Street View environment.

\paragraph{LANI}
\citeauthor{misra2018mapping} proposes LANI where an agent navigates between landmarks, and CHAI where an agent navigates and manipulates objects in the house environment.

\paragraph{DeepMind Lab}
DeepMind Lab environment \cite{beattie2016deepmind} provides a set of first-person 3D mazes for an agent to navigate, and \citeauthor{mirowski2016learning} experiments the effect of different auxiliary losses on the mazes.

\paragraph{Touchdown}
In Touchdown dataset \cite{chen2019touchdown}, in addition to reaching certain locations the agent also needs to use the spatial description to locate the mascot.

Unlike other navigation tasks, IQA \cite{gordon2018iqa} provides agents with questions instead of the instructions.
Agents need to explore the interactive environment to get necessary information to answer questions.
While aforementioned datasets address different aspects of vision and language grounded tasks, ALFRED requires an agent to perform more complicated actions such as navigating via egocentric visual inputs and meanwhile interact with objects by predicting interaction masks.

\subsection{Existing Techniques}
\citeauthor{NEURIPS2018_6a81681a} uses a speaker model to generate instructions from simulated trajectories. \citeauthor{tan-etal-2019-learning} applies the environmental dropout mask to remove some objects from the scenes.

\citeauthor{storks2021we} proposes an approach that integrates an object detection model with a model that predicts the direction of the goal.
\citeauthor{corona2020modularity} uses separate modules to predict actions for different types of instructions. This leads to improved performance on unseen scenes and tasks.
\citeauthor{pratap2020moca} argues that the information for predicting object masks and actions is different (i.e. general instr v.s. step-by-step). By separating the model into object mask prediction and action prediction branches, they achieved the second best on the leaderboard.

\citeauthor{ma2019selfmonitoring} proposes progress estimation that encourages the agent to utilize the instruction sequentially with attention.
Relying on the estimated progress, \citeauthor{ma2019regretful} proposes a regret module that decides whether or not to roll back, and use a progress marker for action selection.
\citeauthor{hu-etal-2019-looking} shows that visual features are not well utilized in unseen scenarios.
They also propose to use Faster-R-CNN~\cite{NIPS2015_14bfa6bb} to construct visual features. It performs better in unseen scenes.

There are also techniques for environments other than ALFRED and Room2Room.
\citeauthor{Cognitive} proposes Cognitive Mapper and Planner, which learns the world map information with a differentiable spatial memory to select actions for an input goal.
\citeauthor{blukis2020few} grounds visual objects with a database for few-shot adaptation in a quadcopter simulator environment.
\citeauthor{andreas-klein-2015-alignment} formulates the instruction to action task as a structure alignment problem using conditional random field (CRF).
Finally, there are some studies that formulate the task as an instruction parsing task.
\citeauthor{karamcheti-etal-2020-learning} designs a pipeline that parses instructions into primitive actions.
\citeauthor{artzi-zettlemoyer-2013-weakly} parses instructions with a combinatory categorial grammars (CCG) parser learned with a weakly supervised algorithm.



\section{Background}

\subsection{Task Formulation}
The ALFRED task~\cite{shridhar2020alfred} aims to learn a mapping from natural language instructions and egocentric vision to sequences of actions for household tasks.
An agent needs to carry out tasks described in given instructions by interacting with an environment.
For each trajectory, at the beginning, a high-level instruction and some step-by-step instructions in natural language are given to the agent.
At each time step, the agent has access to the visual observation from its first person point of view.
The agent needs to output actions according to the instructions and the visual input.
Possible actions include five navigation actions and seven actions that interact with objects in the environment.
To interact with an object in the environment, for example, pick up an object, the model needs to generate a pixel-level mask that selects an object.

\subsection{Dataset}
The ALFRED training dataset consists of trajectories in different environments generated by expert demonstration, each of which is annotated with the corresponding instructions.
We include the statistics about the dataset in Table~\ref{tab:instruction-len}.
The dataset also provides the time span each step-by-step instruction corresponds to.
We will use this additional information in our following analysis and proposed approaches.

\begin{table*}[]
    \centering
    \begin{tabular}{cccc}
    \toprule
        & Train & Valid-Seen & Valid-Unseen \\
    \midrule
    \# of Step-by-Step Instruction & 6.72 (2.49) & 6.79 (2.72) & 6.26
    (2.33) \\
    \# of word in Step-by-Step Instruction & 12.30 (5.93) & 12.13 (6.00) & 12.59 (6.36) \\
    \# of word in High-level Instruction & 10.06 (3.05) & 10.11 (3.17) & 10.05 (2.89) \\
    \bottomrule
    \end{tabular}
    \caption{Statistics of the instructions.}
    \label{tab:instruction-len}
\end{table*}

\subsection{Metrics}
We consider two existing metrics used in ALFRED~\cite{shridhar2020alfred}:
1) \textbf{Task Success}: It is defined as 1 if the task goal-conditions are all met, and 0 otherwise.
2) \textbf{Goal-Condition Success}: The task success is 1 only if all goal-conditions are met, which might be too challenging. To better justify the performance, we report the percentage of completed goal-conditions.
In addition, as shorter trajectories are more efficient and favorable, we also consider the path weighted version of the above two metrics. Concretely, the weight to be multiplied with is calculated as $\frac{L^*}{\max(L^*, \hat{L})}$, where $\hat{L}$ is the number of actions taken by the expert, and $L^*$ is the number of actions taken by the model.

\subsection{Baseline Models}
We choose two publicly available models as our baseline:
(1) The \textbf{Seq2Seq} model proposed in the original ALFRED paper~\cite{shridhar2020alfred}.
Given visual observation, instructions and a goal as inputs, the Seq2Seq model is trained to predict the action sequence with imitation learning.
(2) \textbf{MOCA} \cite{pratap2020moca}. It separates the model into a visual perception module predicting object masks and an action policy module predicting actions.
Since these two predictions require different information, they argued that separating the model into two branches improves the overall performance, achieving the second best performance on the leaderboard when the time we do this work.

Both of the two models use two auxiliary losses.
\begin{enumerate}
    \item A progress monitor (PM) loss:  At each time step $t$, the model predict a progress $\hat{p}_t$. The PM loss is a $l_2$ loss between $\hat{p}_t$ and $p_t$, where $p_t = t / T$ is the progress in terms of the total number of steps $T$.
    \item A sub-goal (sg) loss:  At each time step $t$, the model predict the ratio of completed sub-goals $\hat{c}_t$. The PM loss is a $l_2$ loss between $\hat{c}_t$ and $c_t$, where $c_t = c / C$ is the number of completed sub-goals $c$ over the number of sub-goals in this instance $C$.
\end{enumerate}

\begin{table*}[]
    \small
    \centering
    \begin{tabular}{lccccccc}
    \toprule
        & \multicolumn{2}{c}{Train} & \multicolumn{2}{c}{Seen} & \multicolumn{2}{c}{Unseen} \\
        & Attention & Gradient & Attention & Gradient & Attention & Gradient\\
    \midrule
      random & \multicolumn{2}{c}{.290 (.108)} & \multicolumn{2}{c}{.294 (.115)} & \multicolumn{2}{c}{.328 (.119)} \\
        Seq2Seq~\shortcite{shridhar2020alfred} & .590 (.155) & .594 (.158) & .589 (.153) & .593 (.156) & .354 (.182) & .363 (0=.181) \\
        - p.m. loss &  .575 (.147) & .580 (.149) & .567 (.147) & .571 (.150) & .368 (.191) & .372 (.188) \\
        - subgoal loss & .541 (.146) & .554 (.149) & .544 (.149) & .554 (.149) & .369 (.177) & .382 (.178) \\
        - both & .562 (.152) & .567 (.155) & .551 (.151) & .557 (.155) & .337 (.181) & .342 (.179) \\
        MOCA \shortcite{pratap2020moca}  & .443 (.207) & .382 (.180) & .450 (.208) & .384 (.177) & .436 (.202) & .348 (.165) \\
    \bottomrule
    \end{tabular}
    \caption{Boundary Adherence Score of baselines. A higher score means less violation of alignment. ``Random" is the expected score when the attention score or random score distributes uniformly over the instructions. The metrics on the training set is computed over 820 trajectories randomly sampled from the training data.}
    \label{tab:BAS_baseline}
\end{table*}

\section{Alignment Analysis} \label{sec:alignment}
\paragraph{Motivation}
Intuitively, better alignment between modalities (i.e. visual representations and instructions) should lead to improved performance.
However, previous work~\cite{hu-etal-2019-looking} discovers that models might achieve better performance with the removal of certain modalities, let alone the alignment between them.
While ALFRED~\cite{shridhar2020alfred} justifies the necessity of all modalities for task success, the alignment between them is still not measured explicitly.
Overall, two questions remain unanswered: 1) does the model learn alignment between modalities? 2) If yes, does better alignment lead to task performance improvement? We aim at designing metrics that help us answer the two questions.

\paragraph{Alignment Definition}
We are given two input modalities: a sequence of images, denoted by $\bar{V} = \{v_i\}_{i=1}^T$, and a sequence of words of step-by-step instructions
, denoted by $\bar{S} = \{s_j\}_{j=1}^{L_s}$, where $L_s$ is the word number of instructions and $s_j$ is a word in instructions.
The ground truth alignment $\alpha$ is defined to be a surjective function $f:\bar V \to \bar S$, which is given in the Alfred dataset.

\paragraph{Approaches}
We propose two approaches to identify model alignment $f_M$ between $\bar{V}$ and $\bar{S}$ in a model $\mathcal{M}$.
In the first method, we assume the existence of an attention mechanism between $\bar{V}$ and $\bar{S}$. Let the attention score from $v_t$ to the $k$th token of the $j$th instruction be $\alpha_{t,j}^{(k)}$. We define
\begin{equation}
    f_M( v_t ) = \arg \max_j \max_k \alpha_{t,j}^{(k)}.
\end{equation}


Another way to identify $f_M$ is to inspect the gradient norm of an input instruction.
At each time step $t$, an ALFRED model predicts an action $a_t$ according to the video inputs the model has observed $\{ v_i \}_{i = 1}{t}$.
We calculate the gradient norm of the $k$th token in the $j$th instruction $s^{(k)}_j$ as:
\begin{align}
    g^{(k)}_{t, j} = \left \lVert
        \frac{\partial \mathcal{L}(a_t)}{\partial{\mathrm{Emb}(s^{(k)}_j)}}
    \right \rVert_2,
\end{align}
where $\mathcal{L}(a_t)$ is the loss at time step $t$.
A greater value of gradient norm implies the word contributes more to the action prediction. 
With $g$, we can define 
\begin{equation}
    f_M(v_t) = \arg \max_j \max_k g^{(k)}_{t, j}.
\end{equation}
Note that the action $\{ a_t \}$ serves as a proxy of $V$ here due to the natural mapping between actions and images.

Soft alignment scores such as attention scores or gradient norm can always be transformed to hard alignment function $f_M$ by greedy/max selection.

\paragraph{Boundary Adherence Score}
Finally, we define the \textit{Boundary Adherence Score} (BAS) $B$ as:
\begin{align}
    B  = \frac{1}{L_s}\sum_{i=1}^{L_s} \mathbbm{1}[f(v_i) = f_M(v_i)]
\end{align}
It is the frequency that the model's alignment follows the ground truth alignment.
Obviously, higher $B$ indicates better $f_M$.

\paragraph{BAS of the Baseline Models}

Table~\ref{tab:BAS_baseline} shows the BAS of our two baseline models Seq2Seq and MOCA.
Both the Seq2Seq and MOCA have higher alignment scores than the expected score of random alignment.
This implies that the Seq2Seq model and MOCA are able to align the two modalities to some extent,
Interestingly, MOCA with worse BAS in fact has a far better evaluation result than Seq2Seq.
It shows the alignment ability of models might not be very indicative to the final results.
To verify our motivation that the progress monitor auxiliary loss and the subgoal auxiliary loss are not sufficient for the learning of alignment,  we also apply the intrinsic metrics on the Seq2Seq models trained without using auxiliary losses.
We can see that not using the auxiliary losses affect the alignment metrics by less than 0.04.
It indicates that the auxiliary loss is not very effective to encourage the learning of the alignment.

\section{Models}

\subsection{Proposed Approaches}
\label{sec:propose-approaches}

Based on the analysis of the intrinsic metric, we have found the models fail to align the modalities well.
We also showed adding the progress monitor loss doesn't improve the learning of the alignment.
We conjecture that this maybe be due to two issues
1) The current model architecture cannot utilize the alignment well.
2) Neither progress monitor nor action prediction provides sufficient signal for learning the alignment.

\subsubsection{Neural Program Counter}
\label{sec:program-counter}

To address the first issue, we propose using a \textit{neural program counter} (PC) that adds inductive bias forcing the model to use instructions sequentially.
It is an analogy to the program counter in a CPU,
where a program counter stores the index of the machine code to be executed.
When it is increased by 1, the CPU will execute the next machine code.
Here we treat each low-level step-by-step instruction as an atomic command, and then we define a \textit{soft} neural program counter.
Its value is relaxed to the continuous real number $\mathbb{R}^{+}$.
Specifically, let the value of the neural program counter at the time step $t$ be $c^{(t)}$.
The value is initialized as 0:
\begin{align}
    c^{(0)} = 0.
\end{align}
At each action decoding step $t$, the model can decide whether or not to increase it:
\begin{align}
    c^{(t + 1)} = c^{(t)} + \sigma(f_c(h^{(t)}),
\end{align}
where $f_c$ is a linear layer, $\sigma$ is the sigmoid function, and $h^{(t)}$ is the decoder hidden state at a time step $t$.

Given a sequence of words of step-by-step instructions $\bar{S} = \{s_j\}^{L_s}_{j=1}$, we use $p^{instr}_j$ to denote the index of step-by-step instructions which the word $s_j$ is in.
For example, if the word $s_j$ belongs to the third instruction, then $p^{instr}_j$ is $2$ since the index starts from 0.
In both Seq2Seq baseline model and MOCA model, the decoder computes attention weights $a^{(t)}$ over input instruction words $\bar{S}$ at each decoding step.
To force the model to focus on an specific instruction at each decoding step, we use $c^{(t)}$ to construct an attention mask $m^{(t)} \in \mathbb{R}^{L_s}$ on $a^{(t)}$:
\begin{align}
    m^{(t)}_j = \exp \left\{ -\lambda \left\lvert p^{instr}_j - c^{(t)} \right\rvert \right\},
    \label{eq:attn-mask}
\end{align}
where $m^{(t)}_j$ is the attention mask for the word $s_j$, and $\lambda$ is a parameter to learn whose minimum value is $0$.
The $\lambda$ controls the strictness of the mask.
When $\lambda \to \infty$, $m^{(t)}_j$ becomes a hard 0-1 mask.
On the contrary, when $\lambda \to 0$, $m^{(t)}_j$ is a mask whose all values are equal to 1.
We expect the model can automatically adjust $\lambda$ to make the learning process easier.

The mask $m^{(t)}_i$ is then applied to the original attention weights $a^{(t)}$ to form the new attention weights:
\begin{align}
    \Tilde{a}^{(t)} = a^{(t)} \odot m^{(t)}
    \label{eq:masked-attn}
\end{align}
where $\odot$ denotes the element-wise product.
The model then uses these new attention weights $\Tilde{a}^{(t)}$ to compute the attention output as the original model predicts the action at this step.

\subsubsection{Auxiliary Loss $L_{pc}$}

However, having a program counter itself may not be able to address the second issue mentioned in Section~\ref{sec:propose-approaches}.
The training signal might not be sufficient for the model to learn a proper $c^{(t)}$, which would cause the masked attention $\Tilde{a}^{(t)}$ to be very noisy.
Therefore, we designed an auxiliary loss $L_{pc}$ to make the learning of the proposed program counter easier.
Since the training data provides that ground truth mapping between each action and the corresponding step-by-step instruction, we can compute the oracle program counter value $\bar{c}^{(t)}$ at each time step.
The loss $L_{pc}$ is the Mean Squared Error (MSE) loss between the oracle $\bar{c}^{(t)}$ and the predicted $c^{(t)}$.
\begin{align}
    L_{pc} = \frac{1}{T} \sum_{t=1}^{T} (\bar{c}^{(t)} - c^{(t)})^2
\end{align}
where $T$ the number of time steps in one sample.

\newcommand{\sd}[1]{\scriptsize{#1}}
\begin{table*}[t]
\centering
\begin{small}
\begin{tabular}{@{\hspace{5pt}}l@{\hspace{10pt}}r@{\hspace{5pt}}rr@{\hspace{5pt}}rr@{\hspace{5pt}}rr@{\hspace{5pt}}r@{\hspace{5pt}}}
\toprule
                            & \multicolumn{4}{c}{Valid} & \multicolumn{4}{c}{Test}\\
                            & \multicolumn{2}{c}{Seen} & \multicolumn{2}{c}{Unseen} & \multicolumn{2}{c}{Seen} & \multicolumn{2}{c}{Unseen} \\
Methods                     & Task & G-C & Task & G-C & Task & G-C & Task & G-C \\
\midrule
No Language      & 0.0 \sd{(0.0)} & 5.9 \sd{(3.4)} & 0.0 \sd{(0.0)} & 6.5 \sd{(4.7)} & 0.2 \sd{(0.0)} & 5.0 \sd{(3.2)} & 0.2 \sd{(0.0)} & 6.6 \sd{(4.0)}\\
No Vision        & 0.0 \sd{(0.0)} & 5.7 \sd{(4.7)} & 0.0 \sd{(0.0)} & 6.8 \sd{(6.0)} & 0.0 \sd{(0.0)} & 3.9 \sd{(3.2)} & 0.2 \sd{(0.1)} & 6.6 \sd{(4.6)}\\
Goal-Only        & 0.1 \sd{(0.0)} & 6.5 \sd{(4.3)} & 0.0 \sd{(0.0)} & 6.8 \sd{(5.0)} & 0.1 \sd{(0.1)} & 5.0 \sd{(3.7)} & 0.2 \sd{(0.0)} & 6.9 \sd{(4.4)}\\
Instruction-Only & 2.3 \sd{(1.1)} & 9.4 \sd{(6.1)} & 0.0 \sd{(0.0)} & 7.0 \sd{(4.9)} & 2.7 \sd{(1.4)} & 8.2 \sd{(5.5)} & 0.5 \sd{(0.2)} & 7.2 \sd{(4.6)}\\
\midrule
\newcite{shridhar2020alfred} & 3.7 \sd{(2.1)} & 10.0 \sd{(7.0)} & 0.0 \sd{(0.0)} & 6.9 \sd{(5.1)} & 4.0 \sd{(2.0)} &  9.4 \sd{(6.3)} & 0.4 \sd{(0.1)} & 7.0 \sd{(4.3)}\\
\newcite{storks2021we} & 1.4 \sd{(0.0)} & - & 0.0 \sd{(0.0)} & - & - & - & - & -\\
\newcite{takayuki2020hier} & - & - & - & - & 12.4 \sd{(8.2)} & 20.7 \sd{(18.8)} & 4.5 \sd{(2.2)} & 12.3 \sd{(9.4)} \\
\newcite{pratap2020moca} & 19.2 \sd{(13.6)} & 28.5 \sd{(22.3)} & 3.8 \sd{(2.0)} & 13.4 \sd{(8.3)} & 22.1 \sd{(15.1)} & 28.3 \sd{(22.1)} & 5.3 \sd{(2.7)} & 14.3 \sd{(10.0)}\\
\midrule
MOCA + Oracle PC  & 6.2 \sd{(4.3)} & 14.6 \sd{(12.5)} & 0.5 \sd{(0.2)} & 8.3 \sd{(7.2)} & - & - & - & - \\
MOCA + PC  & 16.6 \sd{(12.1)} & 25.7 \sd{(20.6)} & 1.7 \sd{(0.8)} & 11.7 \sd{(7.9)} & - & - & - & -  \\
MOCA + PC + $L_{pc}$     &    \textbf{19.5 \sd{(14.7)}}  & \textbf{28.9 \sd{(24.0)}}  & 3.9 \textbf{\sd{(2.3)}} & 13.3 \sd{(8.9)} & - & - & - & -\\
MOCA + F.G. PC + $L_{pc}$ & 17.7 \sd{(12.1)} & 25.8 \sd{(20.6)} & \textbf{4.1} \sd{(2.2)} & \textbf{14.2 \sd{(9.1)}} & - & - & - & - \\ 
MOCA + F.G. PC + $L_{pc}$ + $\tau$  & 15.7 \sd{(10.1)}  & 24.0 \sd{(14.9)}  & 3.4 \sd{(1.8)} & 14.0 \sd{(8.6)} & - & - & - & -\\

\midrule
Human             & - & - & - & - & - & - & 91.0 (85.8) & 94.5 (87.6)\\
\bottomrule
\end{tabular}
\end{small}
\caption{Task and Goal Condition Success Rate. For each metric, the corresponding path weighted metrics are given in parentheses. All values are percentages. 
}
\label{tab:baselines}
\end{table*}

\begin{table*}[]
    \small
    \centering
    \begin{tabular}{lccccccc}
    \toprule
        & \multicolumn{2}{c}{Train} & \multicolumn{2}{c}{Seen} & \multicolumn{2}{c}{Unseen} \\
        & Attention & Gradient & Attention & Gradient & Attention & Gradient\\
    \midrule
        random & \multicolumn{2}{c}{.290 (.108)} & \multicolumn{2}{c}{.294 (.115)} & \multicolumn{2}{c}{.328 (.119)} \\
        Seq2Seq \shortcite{shridhar2020alfred}  & .590 (.155) & .594 (.158) & .589 (.153) & .593 (.156) &  .354 (.182) & .363 (.181)  \\
        MOCA \shortcite{pratap2020moca}  & .443(.207) & .382 (.180) & .450 (.208) & .384 (.177) & .436 (.202) & .348 (.165) \\
    \midrule
        MOCA + Oracle PC  & .990 (.033) & .732 (.135) & .989 (.037) & .736 (.136) & .990 (.040) & .718 (.140) \\
        MOCA + PC  & .448 (.135) & .364 (.143) & .429 (.133) & .345 (.144) & .424 (.148) & .336 (.144) \\
        MOCA + PC + $L_{pc}$  & .813 (.121) & .735 (.147) & .777 (.139) & .705 (.155) & .724 (.165) & .646 (.166)\\
        MOCA + F.G. PC + $L_{pc}$  & .566 (.147) & .559 (.155) & .556 (.156) & .550 (.165) & .501 (.147) & .492 (.151) \\
        MOCA + F.G. PC + $L_{pc}$ $\tau$  & .595 (.145) & .545 (.159) & .579 (.158) & .537 (.162) & .534 (.149) & .497 (.154)\\
    \bottomrule
    \end{tabular}
    \caption{Boundary adherence score. A higher score means less violation of alignment. ``Random" is the expected score when the attention score or random score is distributed uniformly over the instructions. The metrics on the training set is computed over 820 trajectories randomly sampled from the training data. 
}
    \label{tab:intrinsic}
\end{table*}

\subsubsection{Fine-grain Program Counter}

We have also observed that there are long instructions which can be further split into multiple shorter sequences.
Since generating attention masks over more fine-grained instructions might allow the model to learn better alignments, we propose to split the long instructions with punctuation and the word "and".
For example, the instruction "Turn to the right and move towards the range, then turn to the right ..." will be split into three short instructions "Turn to the right", "move towards the range", and "then turn to the right ...".
We name the program counter trained on these fine-grained instructions \textit{Fine-grained PC}.

When the fine-grained PC is used, the context of a shorter instruction may become important.
The agent may need to look ahead one instruction in order to know whether it has completed this instruction.
It motivates us to propose to modify Equation~\ref{eq:attn-mask} a bit:
\begin{align}
\begin{split}
    m^{(t)}_j = \exp \{
            &  -\lambda \mathrm{ELU}\left( \left\lvert p^{instr}_j - c^{(t)} \right\rvert - \tau  \right) \\
            &+ \lambda \mathrm{ELU}(-\tau) \}.
    \end{split}
    \label{eq:tau}
\end{align}
Given $c^{(t)}$, $m^{(t)}_j = 1$ if $p^{instr}_j = c^{(t)}$, which is the same as in Equation~\ref{eq:masked-attn}.
However, when $c^{(t)}$ differs $p^{instr}_j$ by less than $\tau$, the attention weights over the $j$th tokens $m^{(t)}_j$ in the instruction will only be less than $1$ by a little.
As a result, the model will be able to attend tokens in the instructions ranging from $c - \tau$ to $c + \tau$.
We use $\tau = 1$ in the following experiments, and name this approach \textit{Fine-grained PC $\tau$}.

\begin{figure*}
\begin{subfigure}{.33\textwidth}
    \centering
    \includegraphics[width=\linewidth]{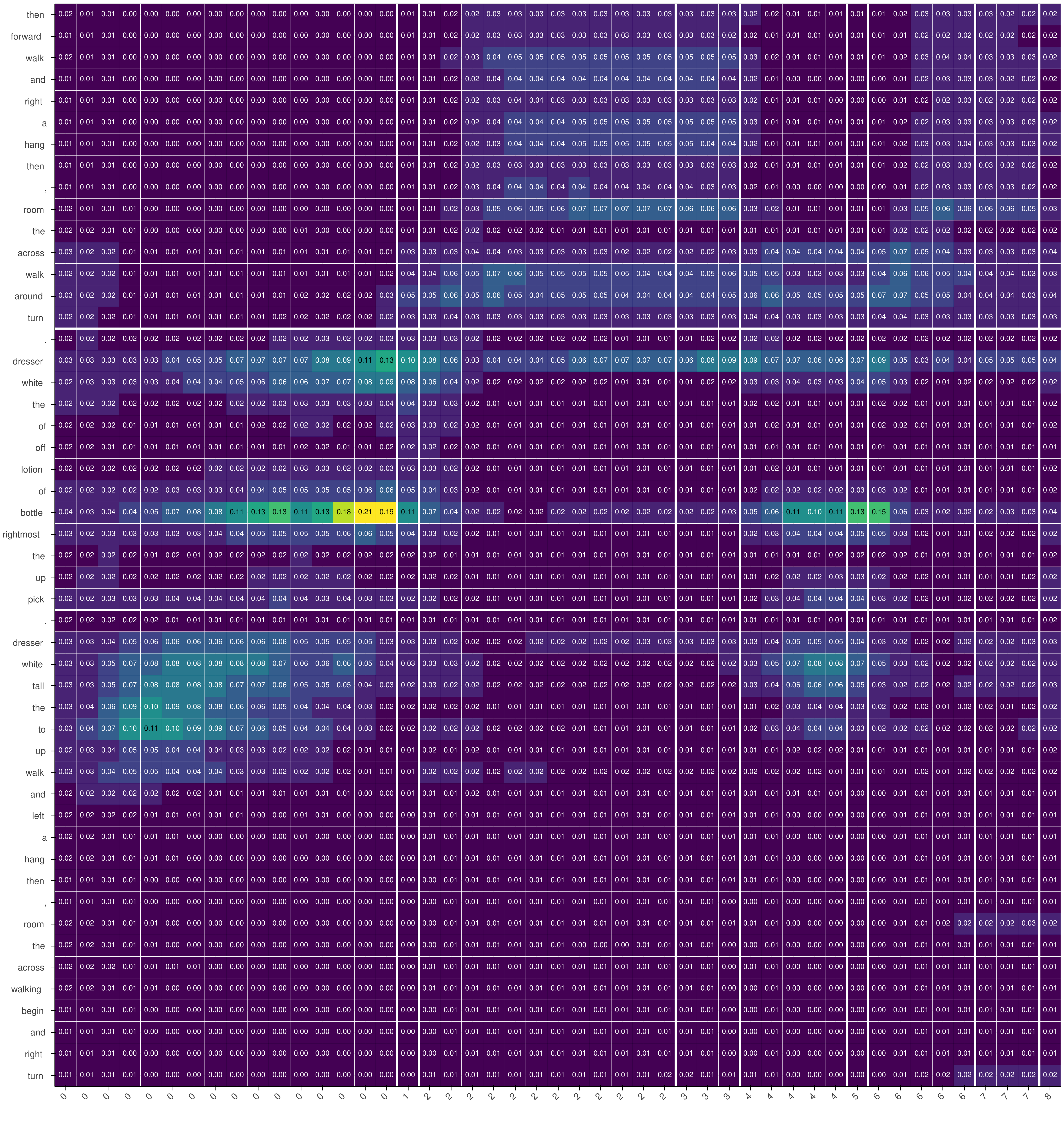}
    \caption{MOCA}
    \label{fig:attn_map_moca}
\end{subfigure}
\begin{subfigure}{.33\textwidth}
    \centering
    \includegraphics[width=\linewidth]{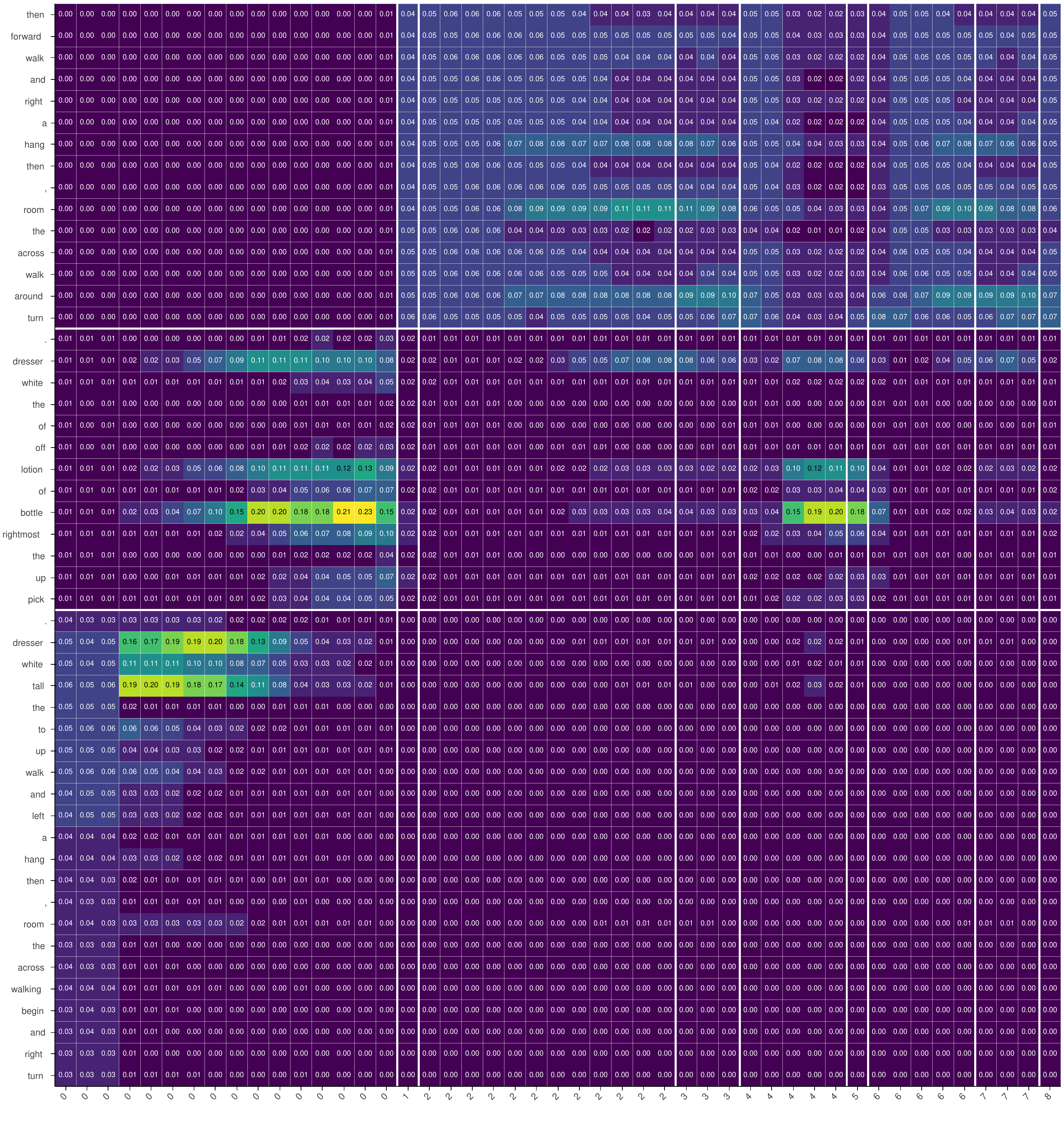}
    \caption{MOCA + PC}
    \label{fig:attn_map_moca_pc}
\end{subfigure}
\begin{subfigure}{.33\textwidth}
    \centering
    \includegraphics[width=\linewidth]{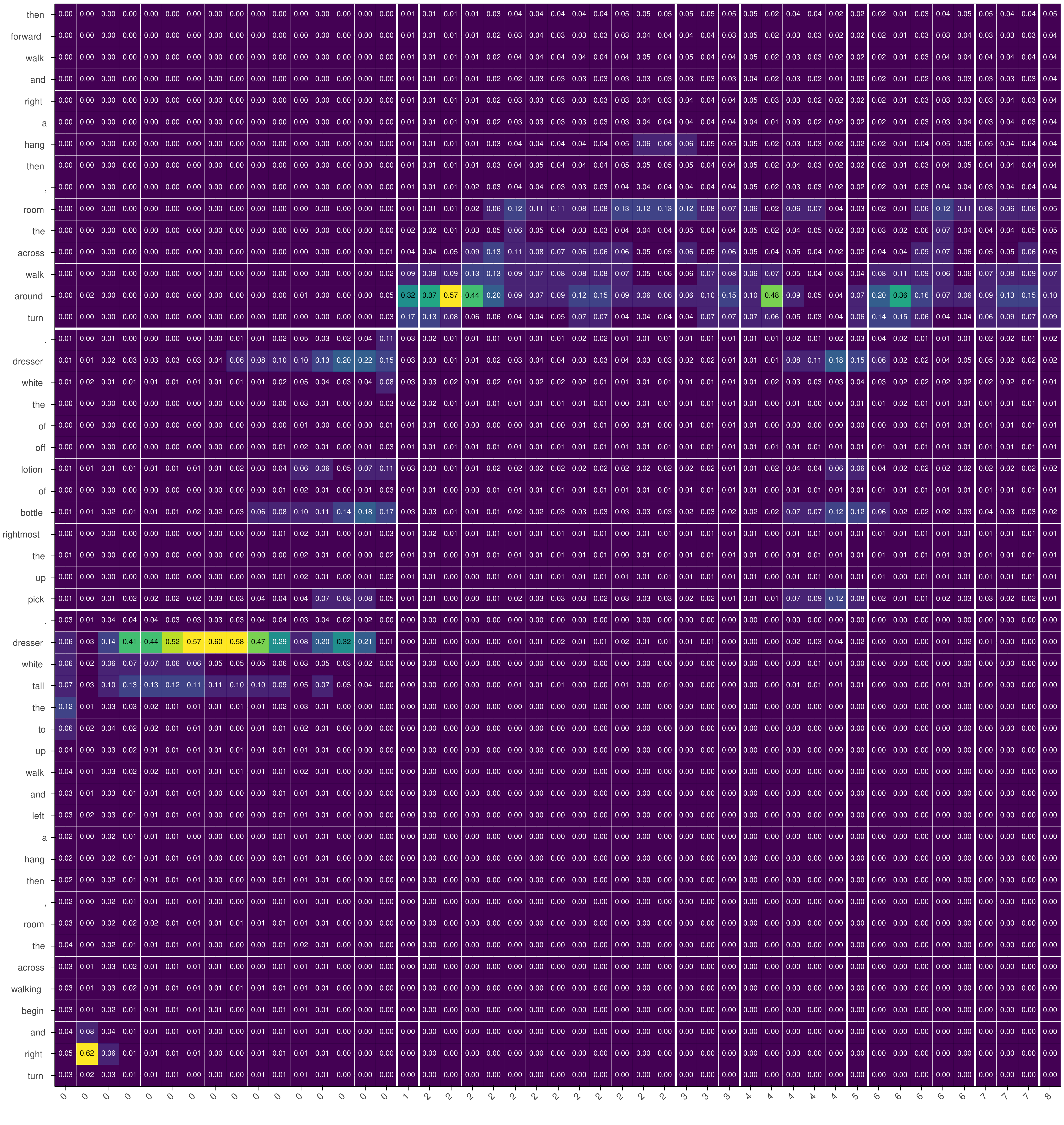}
    \caption{MOCA + PC + $L_{pc}$}
    \label{fig:attn_map_moca_pc_pcloss}
\end{subfigure}
 \caption{Attention map of the MOCA baseline and proposed models. Y-axis is instruction words, and X-axis is the corresponding instruction index of each action. When $L_{pc}$ is used, the attention is more concentrated to the corresponding instruction words. On the other hand, adding program counter alone does not cause much difference compared to the original one.}
 \label{fig:attn_map}
\end{figure*}

\section{Experimental Setup}

We conduct our experiments on the ALFRED dataset following the setting in the \cite{pratap2020moca}.
Due to the constraint of computing resources, we train our models with fewer update steps.
When the auxiliary loss $L_{pc}$ is used, we weight it by 0.2.

\section{Results and Discussion}






\subsection{Neural Program Counter} \label{sec:results_pc}

The results are shown in Table~\ref{tab:baselines}.
When the program counter and $L_{pc}$ are used, our model is able to outperform both the reported and reproduced MOCA models on seen and unseen environments.
The proposed model has larger performance gain on the path weighted metrics, which indicates the model with PC and $L_{pc}$ does less redundant actions.
It shows the superiority of the proposed methods on helping the model learn to follow instructions.

When $L_{pc}$ is not used, the performance is worse.
Especially, compared to MOCA, the task success rate in the unseen settings drop from 3.0 to 1.7.
It is aligned with our hypothesis that the alignment between the instruction and the other modalities is hard to learn by using the original training objective alone.
On the other hand, though the fine-grained program counter cannot improve the performance in the seen setting, when $\tau$ is used, it is able to boost the performance in the unseen setting, achieving the best results on the goal success rate.
It suggests that the program counter and $L_{pc}$ could be more important in the unseen setting, and the fine-grain program counter could possibly further improve the model's generalization capability.

\subsection{Boundary Adherence Score}



Here, we analyze the proposed methods with BAS and show results in Table~\ref{tab:intrinsic}.
From Table~\ref{tab:intrinsic}, we can see both the attention-based and the gradient-based scores are higher when our $L_{pc}$ is used.
This matches our motivation that better alignment can improve the performance.
On the other hand, when the  fine-grained program counter is used, the adherence score is lower than the coarse-grained ones.
One possible explanation is that errors accumulate more severely since there are more steps in the fine-grain setting.

Another interesting observation is that when the program counter is used, the gradient-based scores differ from the attention-based scores more.
Especially, when the oracle counter is used, even though the attention-based score is nearly perfect, the gradient-based scores are only around 0.70.
It suggests that some information across the instruction boundaries is carried by the LSTM encoder layer, and such information may be essential to the model.

\subsection{Qualitative Analysis and Examples}

To further understand how the proposed program counter works, we do qualitative analysis by drawing the attention map in Figure~\ref{fig:attn_map}.
We simply use the first sample in the valid-seen split of the data to draw the figures.
The Figure~\ref{fig:attn_map_moca} shows the attention map of the vanilla MOCA model.
We can observe that the MOCA model already has some extent of the ability to align instructions and actions without any modification, which is coherent with the analysis of the proposed BAS on baselines in Section~\ref{sec:alignment}.
The Figures~\ref{fig:attn_map_moca_pc} and \ref{fig:attn_map_moca_pc_pcloss} are the attention maps of the MOCA + PC model and the MOCA + PC + $L_{pc}$ model respectively.
Compared to the vanilla MOCA model, the attention distributions generated by the models with PC and $L_{pc}$ are more concentrated to the corresponding instruction words.
On the other hand, there is no very significant difference between the attention maps of MOCA model and MOCA + PC model.
Therefore, we can draw a similar conclusion to Section~\ref{sec:results_pc} that it is hard to learn the alignment between modalities by using only the original training objective.

\section{Future Work}
It is promising to incorporate our program counter module and $L_{pc}$ in the very recently proposed models.
For example, our approach can be used in \citet{blukis2021persistent} for leveraging the information from step-by-step instructions.
They can also be used in \citet{kim2021agent} as a replacement of the attended visual features.
\citet{zhang2021hierarchical} executes the step-by-step instructions in order explicitly.
It will be interesting to analysis their alignment with our proposed intrinsic metrics.
We leave the explorations for future work.

\section{Conclusion}

In this work, we investigate the model's ability to align different modalities in the vision and language navigation task ALFRED.
With the proposed Boundary Adherence Score, we find the existing models Seq2Seq and MOCA fail to align the instruction well with other modalities when predicting the action.
Therefore, we further propose the \textit{Neural Program Counter} and the auxiliary loss $L_{pc}$ to help the model learn better multi-modal alignments.
With our proposed methods, we can outperform the state-of-the-art released model MOCA.

To sum up, our contributions include
1) We propose the intrinsic metric BAS score, which can serve as an analysis tool for modality alignment.
2) We discover that previous models do not align the modalities well.
3) We propose the program counter model as well as the $L_{pc}$ auxiliary loss, and outperform the previous strong baseline.

\section*{Acknowledgments}
We would like to thank Yonatan Bisk for his in-depth discussions. We are also thankful to the anonymous reviewers for their comments on the paper.

\bibliography{anthology,custom}
\bibliographystyle{acl_natbib}

\end{document}